\documentclass{article}
\usepackage[nonatbib, final]{neurips_2024}
\usepackage{hyperref}
\usepackage{float}
\usepackage{graphicx} % Required for inserting images
\usepackage{xcolor}
\usepackage{colortbl}
\usepackage{listings}
\usepackage{booktabs}
\usepackage{amsmath}
\usepackage{float}
\usepackage{biblatex}
\usepackage{threeparttable}
\usepackage{amssymb}

\title{Developing Training Procedures for
Piecewise-linear Spline Activation Functions in Neural
Networks}

\author{%
  William H Patty \\ \\
  \textit{An extension of thesis research advised by Michael} \\ 
  \textit{Linderman at Middlebury College} \\ \\
  September 2025
}

\begin{document}
\maketitle

\begin{abstract}

Activation functions in neural networks are typically selected from a set of empirically validated, commonly used static functions such as ReLU, tanh, or sigmoid. However, by optimizing the shapes of a network's activation functions, we can train models that are more parameter-efficient and accurate by assigning more optimal activations to the neurons. In this paper, I present and compare 9 training methodologies to explore dual-optimization dynamics in neural networks with parameterized linear B-spline activation functions. The experiments realize up to 94\% lower end model error rates in FNNs and 51\% lower rates in CNNs compared to traditional ReLU-based models. These gains come at the cost of additional development and training complexity as well as end model latency.
\end{abstract}

\section{Introduction}

\subsection{Activation Functions}
Linear functions cannot model nonlinear correlations and any linear model, irrespective of size, can be approximated by a single layer of a neural network, effectively bounding the abstraction depth of linear models at 1 \cite{Datta2020, Bishop2006}. Nonlinear models, however, can capture both linear and nonlinear relations, and neural networks can be nonlinear if they incorporate nonlinear activation functions \cite{Datta2020}. With the right non-polynomial activations, a feed-forward neural network can theoretically approximate any continuous function \cite{Augustine2024}. However, there is no universally best activation as the effectiveness of function assignments varies depending on the task and even across different sub-components of a model \cite{Szandata2021, Bohra2020, Scardapane2016, Trentin2001}. Each activation comes with its own respective advantages and disadvantages \cite{Szandata2021}. For example, the sigmoid function performs well in binary classification, tanh is particularly useful in recurrent neural networks, and ReLU is usually the first-choice for hidden layers \cite{Szandata2021}.

\vspace{10pt}

When training neural networks, a model's parameters are optimized to its fixed architecture, which includes any preselected activation functions. At model initialization, the post-training roles of the neurons are yet to be determined, as those roles are dependent the training process and which end minima the network converges to in the optimization space. As such, manually preselecting activation functions must be inherently suboptimal. Learnable activation functions, however, can be optimized alongside other parameters such that each function is tuned to the specific role of its corresponding neuron \cite{Bohra2020, Scardapane2016, Trentin2001}. The goal is to train specialized activation functions that maximize each neuron's representational power to train more accurate parameter-efficient models. This architecture change can extend non-linear mapping capabilities, model flexibility, unlock improved model performances that are not reliably replicable by fixed-activation models \cite{Trentin2001, Chen1996}.

\subsection{B-splines}
\begin{figure}[H]
    \centering
    \caption{Example Linear B-spline with Basis Functions}
    \label{fig:bspline_w_basis}
    \includegraphics[scale=0.41]{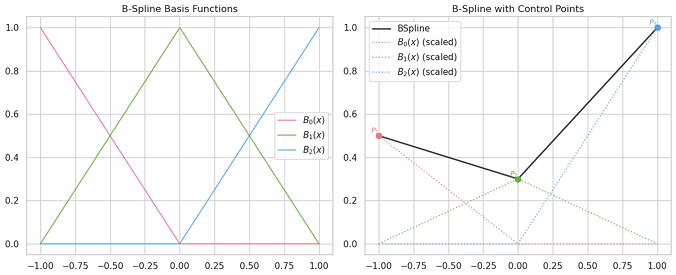}
\end{figure}

Basis spines, or B-splines, are piecewise polynomial curves constructed to fit specified control points. B-splines are defined by basis functions: $d$-dimensional polynomials associated with control points on the spline. Basis functions determine the influence of the control points at each location on the curve. As visualized in Figure \ref{fig:bspline_w_basis}, a B-spline curve's value at every point in the domain is defined as the sum of the coefficients of its control points weighted by the value of their corresponding basis functions at that position.

Control point locations determine the shape of each segment's curvature and the piecewise segments are joined at knots, so by adjusting control point locations, the shape of a B-spline curve can be modified. By mathematical design, degree $d$ B-splines are expressly continuous to their $(d-1)^{th}$ derivative. This property is relevant within a broader discussion of the potential optimality of B-splines as activation functions and motivated their selection for this work, however, this research was conducted in a compute and time constrained environment so the B-splines implemented and discussed here are linear.

\section{Related Works}
\subsection{Parametrization of Existing Activations}

One simple approach to training activation functions is to introduce learnable parameters into an existing activation function to enable optimization by scaling specific pieces of a known function. For example, training a property such as amplitude effectively allows model optimizers to traverse the original parameter search space along another axis by scaling neuron outputs without needing to modify connection weights \cite{Trentin2001}.

\begin{figure}[h!]
    \centering
    Example Parameterized Activation Functions
    \\[2ex]
    \vspace{1ex}
    \begin{minipage}{0.45\textwidth}
        \centering
        (a) Leaky ReLU
        \[
        \text{LReLU}(x) = 
        \begin{cases}
        x, & x \geq 0 \\[6pt]
        \alpha x, & x < 0
        \end{cases}
        \]
    \end{minipage}\hfill
    \begin{minipage}{0.45\textwidth}
        (b) Scaled Sigmoid
        \vspace{1ex}
        \centering
        \[
        f(x) = \frac{a}{1 + e^{-b(x - c)}}
        \]
        
    \end{minipage}
    \label{fig:param-activs}
\end{figure}

It has been shown that models with fixed activation functions can't reliably reach the same minima as those with trainable activations \cite{Trentin2001}. Reductions in end-model error, convergence times, and overfitting have been shown in application domains such as function fitting, audio classification, audio processing, computer vision, signal processing, and translation tasks via parameterizations of the sigmoid, ReLU, and hyperbolic tangent functions \cite{Chen1996, Trentin2001, Goh2003, he2015}. The authors of SWISH, a parameterized, modified sigmoidal function, also noted that SWISH boasted additional gains on smaller models, aligning well with the theory that training activations can lead to more efficient per-neuron and per-parameter performance \cite{Ramachandran2017}.  

\subsection{Combining Activation Functions}

Each additional parameter in a trainable activation function can increase its structural flexibility, and with that the dimensional complexity of the optimization search space. By integrating more flexible activations, we can theoretically open up new paths to reach deeper minima. Consider an approach involving the learned selection of activations from a static set: a change in selection effectively represents a jump from one fixed-activation search space to another. However, if we represent this selection as a weighted combination, we smoothly merge the disjoint search spaces in a gradient-pathable way. Learning weighted combinations of activations has shown improved performance over static activations in reinforcement learning, function fitting, and classification tasks \cite{Manessi2018, Turner2014}. Turner et al. also experimented with parameterizing functions within a weighted combination, but found that the combined approach did not improve model performance, alluding to the risks of over-fitting with hyper-flexible models \cite{Turner2014}. 

\subsection{Polynomial and Hybrid Activations}

In further pursuit of activation flexibility, researchers have explored fully custom, trainable polynomial and hybrid activation functions. Learnable polynomial functions have shown end-loss improvements over static activations on non-linear system identification, function fitting, and computer vision tasks \cite{Piazza1992, goyal2020}. However as adaptable as polynomial functions may be, they are unbounded and non-squashing, which generates an increased risk of unconstrained variability in terms of output scaling within networks that can have consequences in both forward input processing and gradient computation \cite{Vecci1998, Campolucci1996}. Trainable hybrid activations, with both linear and sigmoidal components, have also shown improvements over static activations on image classification tasks \cite{Bodyanskiy2022, Bodyanskiy2023}. However, like polynomial approaches, hybrid activations do not have local control, which means that the optimizer can't modify specific parts of the function shape without changing the rest: any adjustments affect the function globally across its full domain \cite{Scardapane2016}.

\subsection{Learning Piecewise Activations}

Some piecewise functions, however, do exhibit local control, which is highly desirable and requisite of any theoretically optimal activation. Functions with local control can be finely optimized across specific input regions \cite{Aziznejad2020}. Trainable linear piecewise functions have shown performance improvements in image classification and physics modeling compared to static activations \cite{Agostinelli2015}. As well, the cubic Catmull-Rom spline was shown to outperform static functions as an activation for non-linear signal identification, regression, function-fitting, and noisy character recognition tasks \cite{Campolucci1996, Scardapane2016, Vecci1998}. B-splines, smoother extensions of Catmull-Rom splines, have shown improvements in binary classification, image classification, graph node classification, mesh shape correspondence, area classification, and signal recovery \cite{Yu2024, Fey2018, Bohra2020}.

\section{Methods}

\subsection{Training Methodologies}

In this work, I explore and compare 9 different training strategies to meaningfully investigate dual-optimization training dynamics of connection weights and linear B-spline activation functions. I consider sequences of the following methods:

\begin{enumerate}
    \item R (ReLU) or FB (Frozen B-spline): Training the connection weights and biases with fixed activation functions.
    \item WB (Weighted B-spline): Training connection weights and biases simultaneously with activation functions.
    \item B (Isolated B-spline): Training the activation functions exclusively without optimizing connection weights and biases.
    \item WBlrs: Training activation functions alongside the connection weights and biases, which are on a learning rate scheduler. Over the course of training, we decrease the magnitude of the updates made to the connection parameters until we’re only training the activations.
\end{enumerate}

The sequences tested are: 
\begin{enumerate}
    \item R
    \item R to B
    \item R to WB
    \item R to WBlrs
    \item WB
    \item WB to B
    \item WB to FB
    \item WB to WBlrs
    \item WBlrs
\end{enumerate}

All trainable activation functions are initialized in a ReLU shape. 

\subsection{Dataset specifics}
I explore model comparisons across four task categories: regression, function fitting, image classification, and audio classification. The datasets tested are: \href{https://scikit-learn.org/stable/modules/generated/sklearn.datasets.fetch_california_housing.html}{California Housing}, \href{https://archive.ics.uci.edu/dataset/109/wine}{Wine Quality (red only)}, Two Spiral, an arbitrary function: np.tanh(x) * np.cos(y**2) + np.sin(x * y), \href{https://docs.pytorch.org/vision/main/generated/torchvision.datasets.MNIST.html}{MNIST}, \href{https://docs.pytorch.org/vision/main/generated/torchvision.datasets.CIFAR10.html}{CIFAR10}, \href{https://github.com/karolpiczak/ESC-50}{ESC50}, and 
\href{https://docs.pytorch.org/tutorials/intermediate/speech_command_classification_with_torchaudio_tutorial.html}{Speech Commands}. Each dataset is divided into an 85/15 training-validation split for each run. Results presented are validation loss, averaged at each epoch across the median 80\% of runs. For each dataset, hyper-parameters were tested and selected in the following order: number of epochs for convergence, number of runs for consistency, learning rates (connection weights then activation functions), batch size, training change epoch, resample rate (audio datasets only), and class limits (audio datasets only). Three control points were used in each spline and a kernel size of three was used for the image and audio datasets.

\subsection{Implementation reference}
My work implements the \href{https://github.com/joaquimcampos/DeepSplines}{DeepSplines repository} for the linear B-spline activation functions. The codebase for these experiments is available \href{https://github.com/liam-hp/deepsplines-thesis}{here}.

\section{Results}

In section \ref{sect:posttraining}, I compare post-training methods to traditional ReLU training to explore how much potential further improvement trainable activations introduce when swapped into pretrained models to replace static activations. The graphs display averaged loss over epochs for training sequences 1-4.

In section \ref{sect:parameq}, I visualize model-level results showing final losses, calculated by averaging over the last 5 epochs of training, for models trained using each of the specified methodologies at varying parameter counts.

\subsection{Post-training results}
\label{sect:posttraining}
\begin{figure}[H]
    \centering
    \begin{minipage}{0.48\textwidth}
        \centering
        \includegraphics[width=\linewidth]{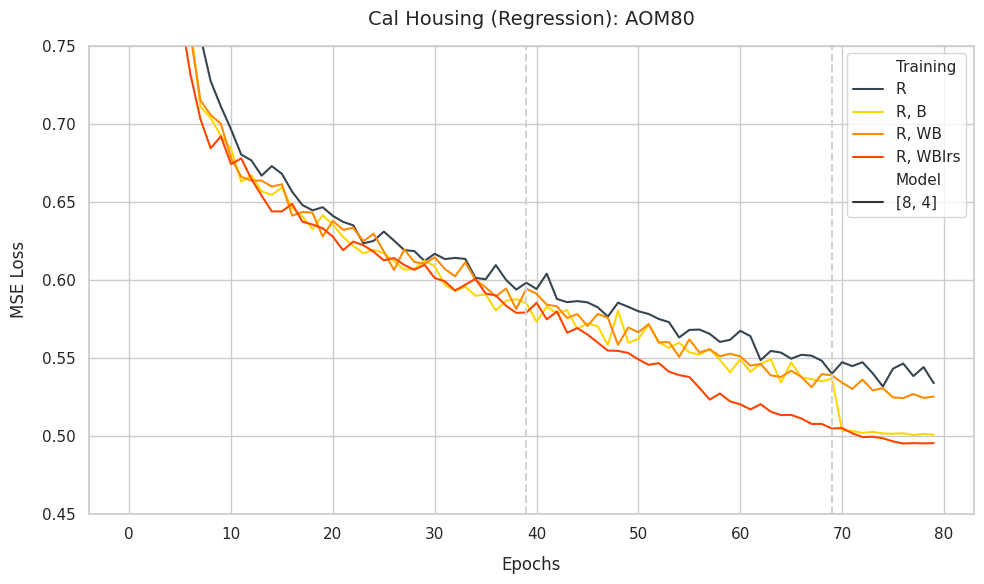}
        \label{fig:calpost}
    \end{minipage}\hfill
    \begin{minipage}{0.48\textwidth}
        \centering
        \includegraphics[width=\linewidth]{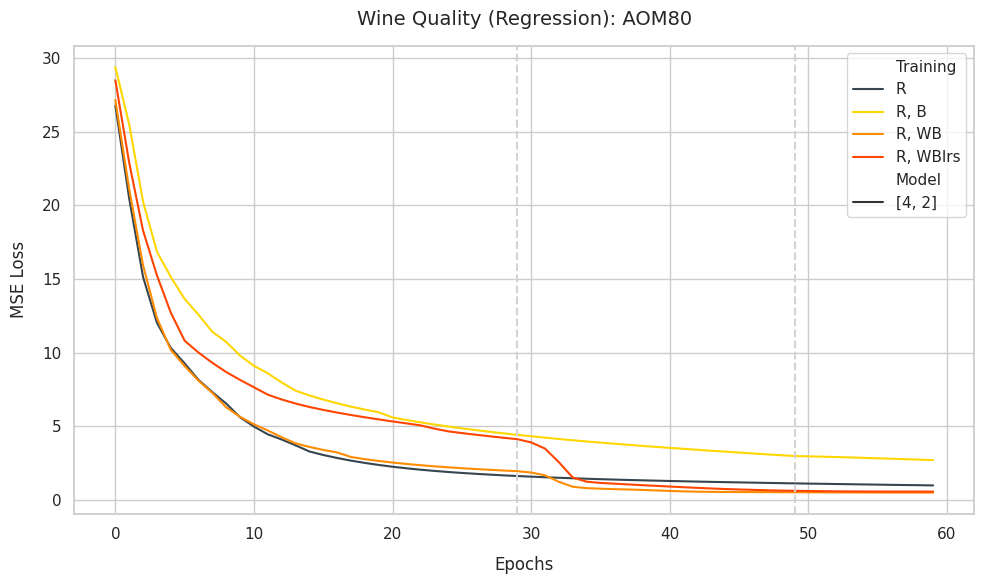}
        \label{fig:winepost}
    \end{minipage}
\end{figure}

\begin{figure}
    \centering
    \begin{minipage}{0.48\textwidth}
        \centering
        \includegraphics[width=\linewidth]{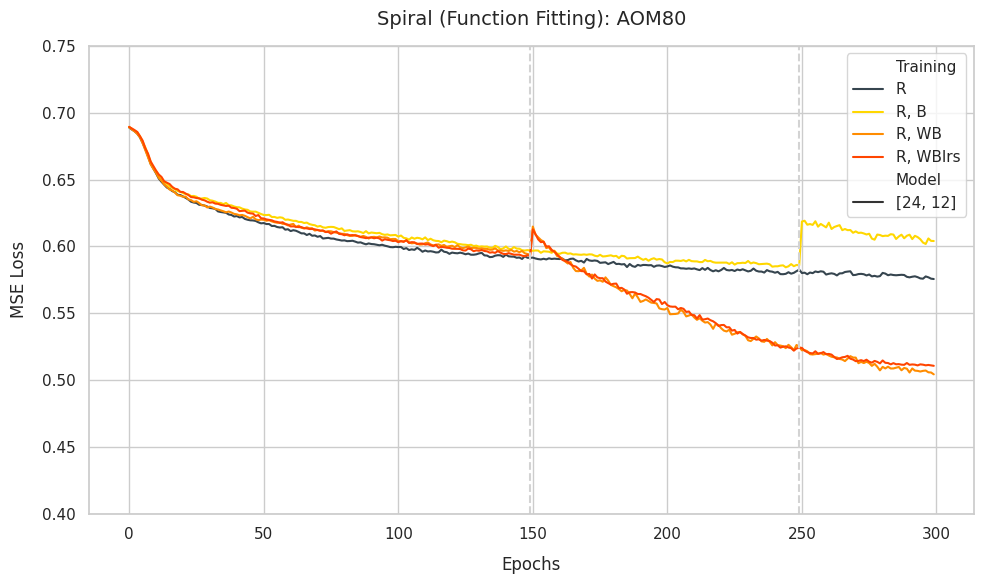}
        \label{fig:spiralpost}
    \end{minipage}\hfill
    \begin{minipage}{0.48\textwidth}
        \centering
        \includegraphics[width=\linewidth]{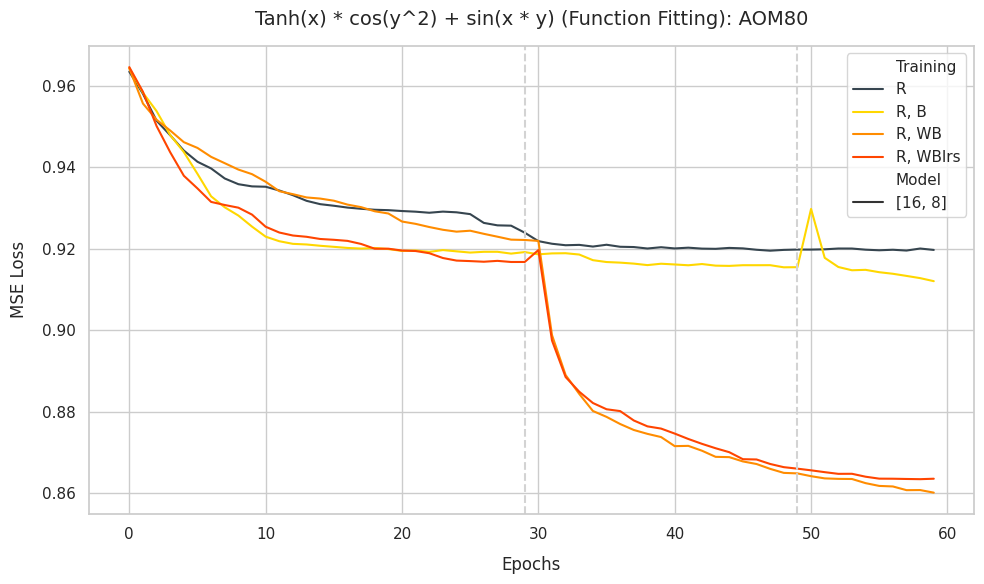}
        \label{fig:funcpost}
    \end{minipage}
\end{figure}

\begin{figure}
    \centering
    \begin{minipage}{0.48\textwidth}
        \centering
        \includegraphics[width=\linewidth]{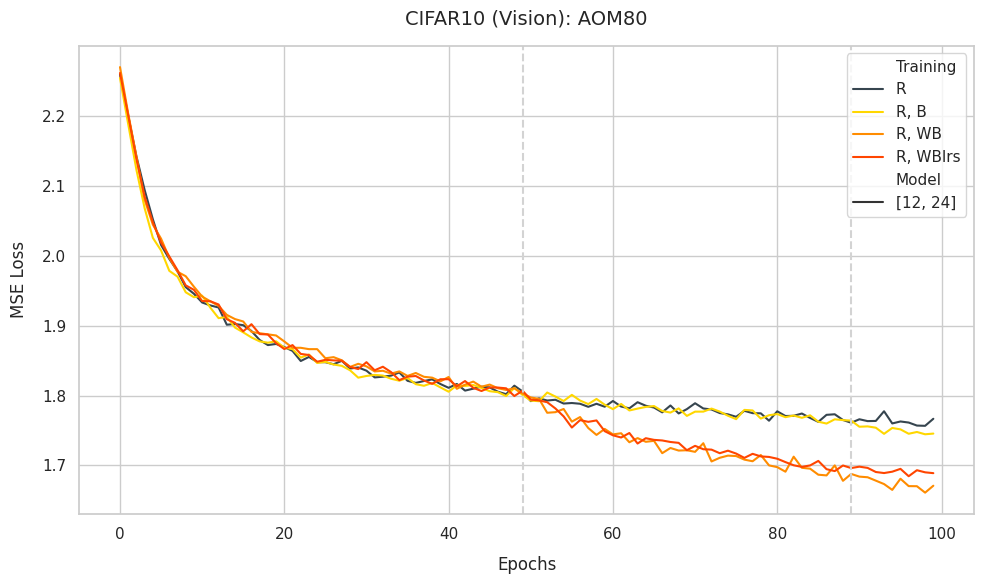}
        \label{fig:cifarpost}
    \end{minipage}\hfill
    \begin{minipage}{0.48\textwidth}
        \centering
        \includegraphics[width=\linewidth]{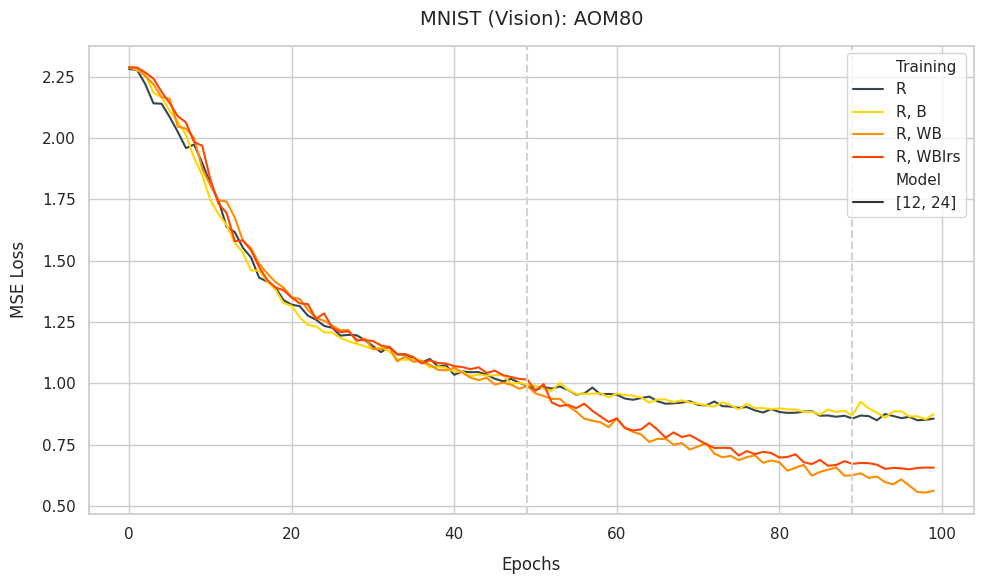}
        \label{fig:mnistpost}
    \end{minipage}
\end{figure}

\begin{figure}[H]
    \centering
    \begin{minipage}{0.48\textwidth}
        \centering
        \includegraphics[width=\linewidth]{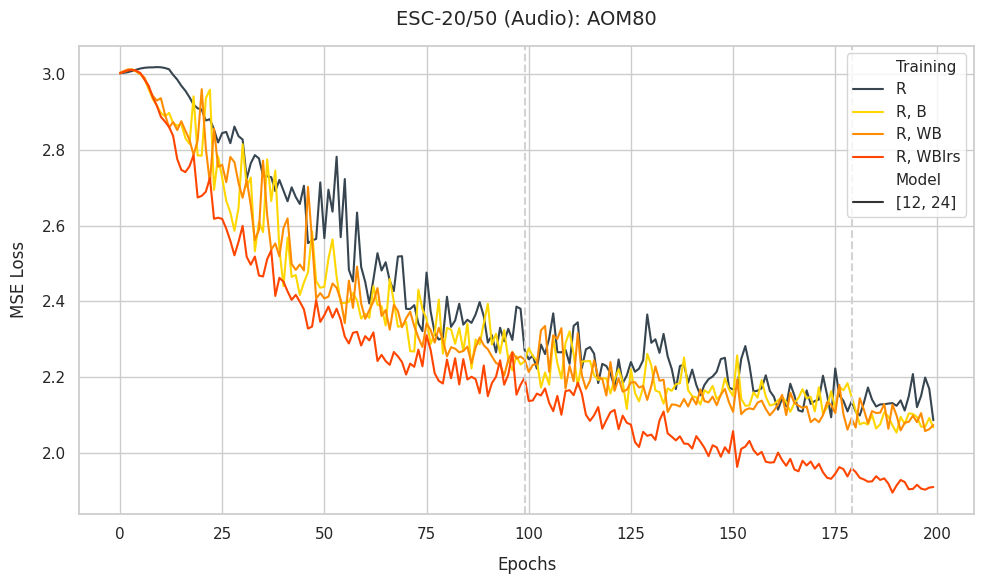}
        \label{fig:escpost}
    \end{minipage}\hfill
    \begin{minipage}{0.48\textwidth}
        \centering
        \includegraphics[width=\linewidth]{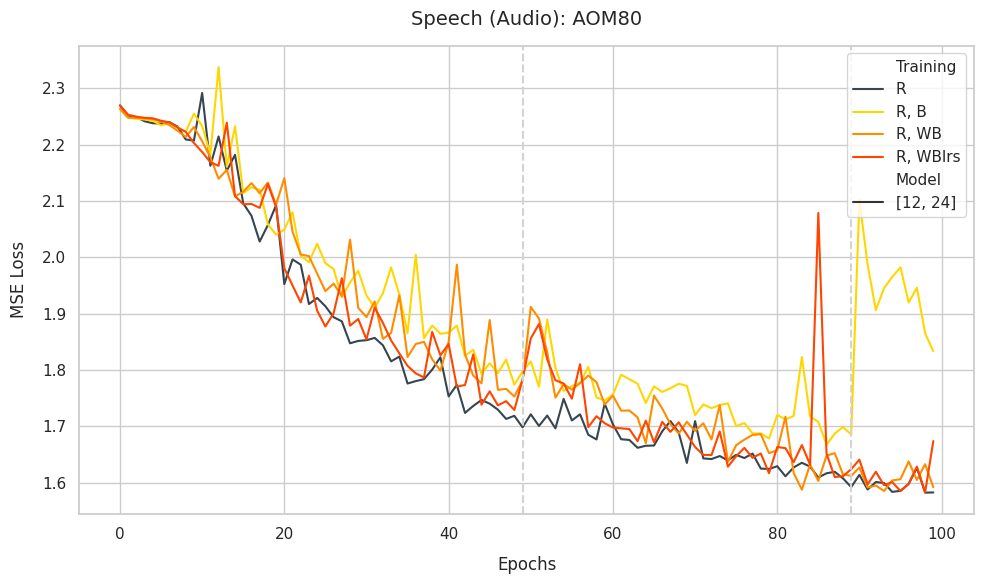}
        \label{fig:speechpost}
    \end{minipage}
\end{figure}

These results reveal a trend of additional subsequent model learning and loss reduction after the static ReLU activations are replaced by trainable splines. Average end loss reductions on optimal model size for each dataset ranged from -0.65\% to 57.93\% with an average improvement across all datasets and tasks of 15.13\%.

\subsection{Parameter equivalent model comparisons}
\label{sect:parameq}
\begin{figure}[H]
    \centering
    \begin{minipage}{0.48\textwidth}
        \centering
        \includegraphics[width=\linewidth]{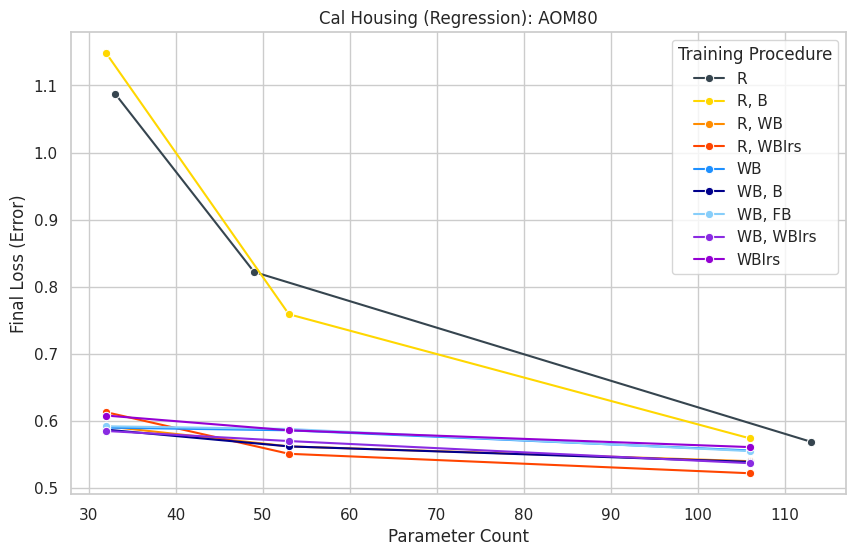}
        \label{fig:calpost}
    \end{minipage}\hfill
    \begin{minipage}{0.48\textwidth}
        \centering
        \includegraphics[width=\linewidth]{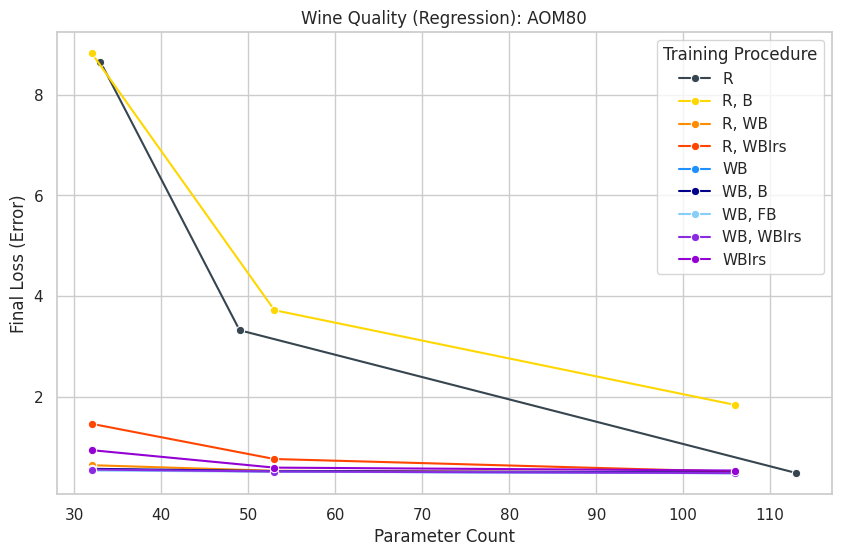}
        \label{fig:winepost}
    \end{minipage}
\end{figure}

\begin{figure}[H]
    \centering
    \begin{minipage}{0.48\textwidth}
        \centering
        \includegraphics[width=\linewidth]{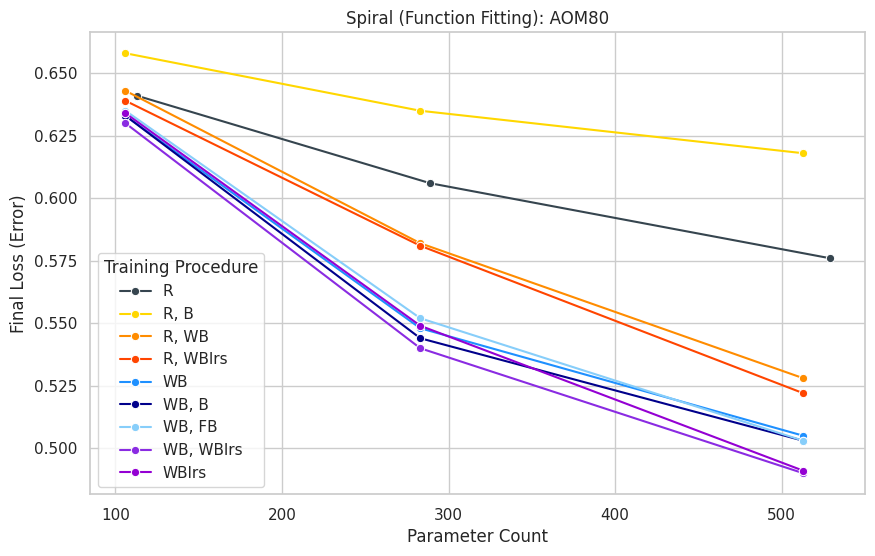}
        \label{fig:spiralall}
    \end{minipage}\hfill
    \begin{minipage}{0.48\textwidth}
        \centering
        \includegraphics[width=\linewidth]{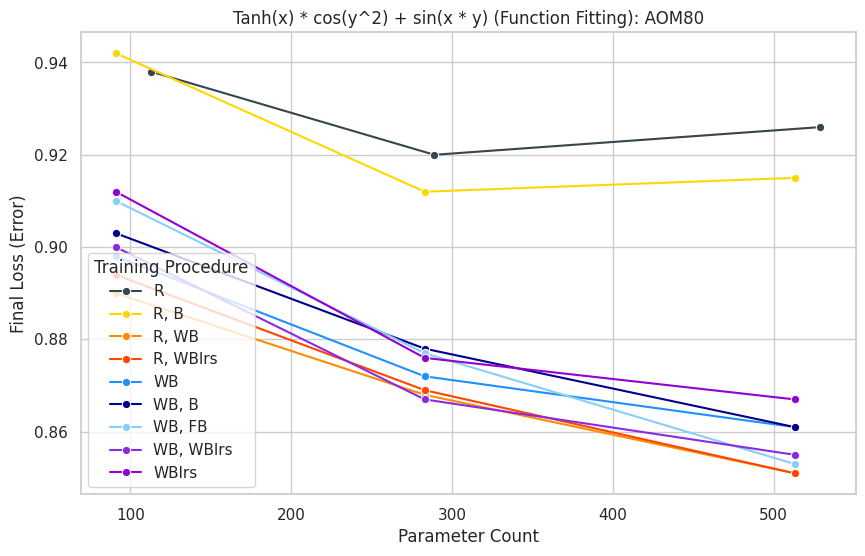}
        \label{fig:funcall}
    \end{minipage}
\end{figure}

\begin{figure}[H]
    \centering
    \begin{minipage}{0.48\textwidth}
        \centering
        \includegraphics[width=\linewidth]{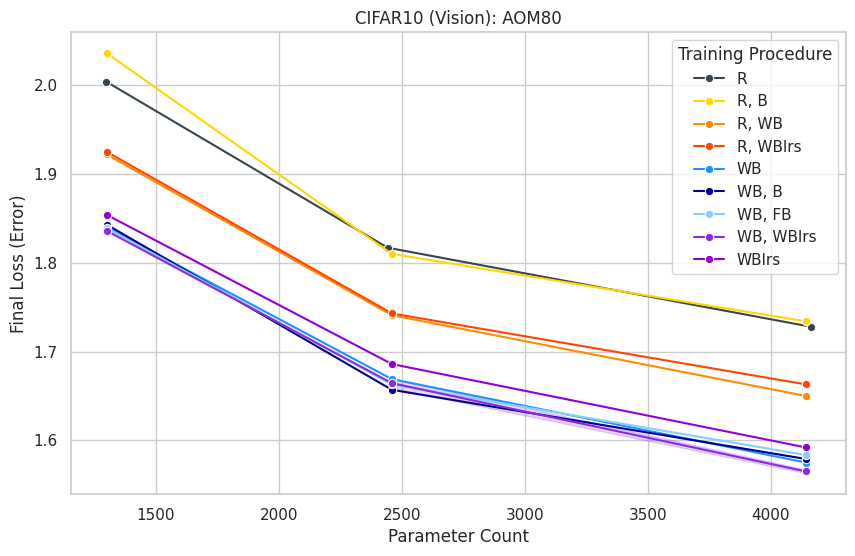}
        \label{fig:cifarall}
    \end{minipage}\hfill
    \begin{minipage}{0.48\textwidth}
        \centering
        \includegraphics[width=\linewidth]{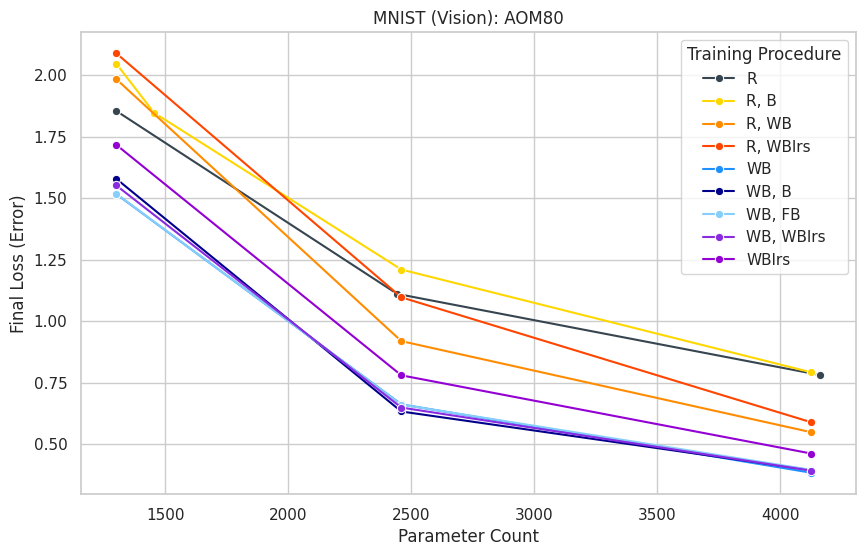}
        \label{fig:mnistall}
    \end{minipage}
\end{figure}

\begin{figure}[H]
    \centering
    \begin{minipage}{0.48\textwidth}
        \centering
        \includegraphics[width=\linewidth]{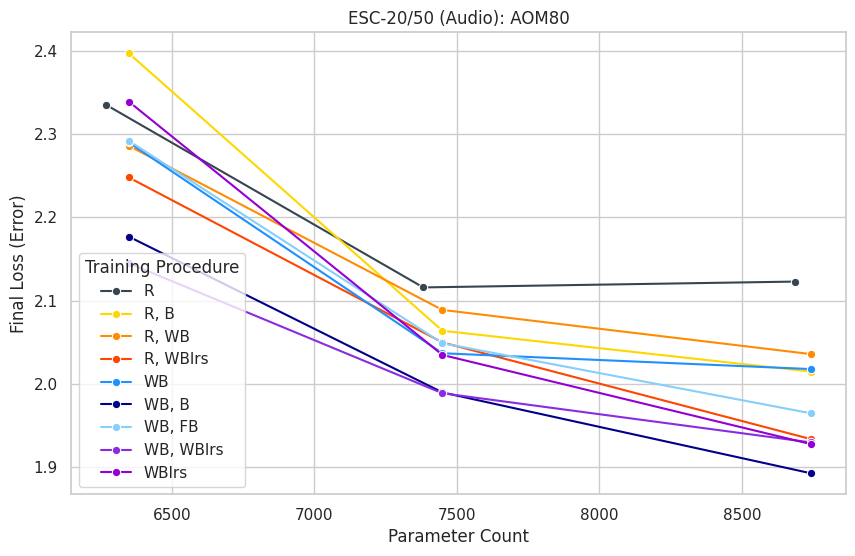}
        \label{fig:escall}
    \end{minipage}\hfill
    \begin{minipage}{0.48\textwidth}
        \centering
        \includegraphics[width=\linewidth]{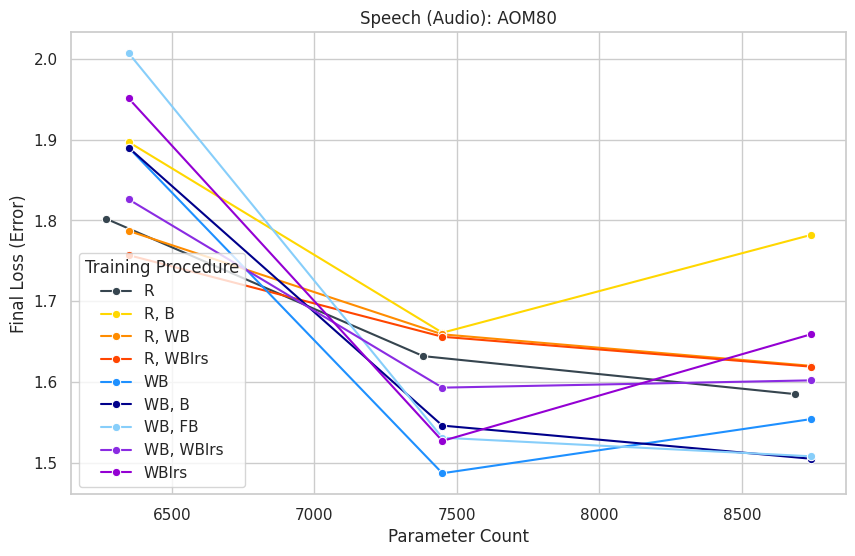}
        \label{fig:speechall}
    \end{minipage}
\end{figure}

End-model comparisons vary by dataset but a trend showing end loss improvements in B-spline based models compared to ReLU is clearly visible. Exemplified by performance on the regression datasets, parameterized B-spline activation functions show potential for training significantly smaller models to reach errors rates that are at least as low as much larger ReLU models. Performance on the function fitting task indicates that B-spline based models can even reach ReLU-unattainable losses, i.e. significantly lower error rates which the ReLU models cannot appear to match, irrespective of model size and training duration.

The only anomalous result in this series of experiments is performance on the speech dataset. This is possibly a factor of the limited time and compute resources available for this research as translated to suboptimal hyper-parameters and data processing. Regardless, it is evidence that training models with learnable spline activation functions needs to be done carefully and consciously. The benefits of these trainable activations appears to vary on a task-to-task as well as a dataset-to-dataset basis, and the model architecture change cannot be reliably drag and dropped in.

In general, we see that the models pretrained on weighted B-spline architectures tended to outperform those pretrained on ReLU architectures. WB to WBlrs was the most consistently dominant procedure in terms of final model performance. ReLU to isolated B-spline showed consistently weaker improvements or even worse results relative to base ReLU. Table \ref{tab:improvements} displays end model performance by task category over the 8 non-ReLU training sequences via final loss reduction compared to a parameter equivalent ReLU model.

\begin{table}[H]
    \centering
    \caption{Loss Reduction by Dataset (AOM80)}
    \label{tab:improvements}
    \begin{tabular}{|c|c|l|c|c|}
        \hline
        \textbf{Task} & \textbf{Dataset} & \textbf{Best} & \textbf{Avg} \\
        \hline
        Regression & California Housing & 46.23\% & 23.39\% \\
        \hline
        Regression & Wine Quality & 93.87\% & 37.90\% \\
        \hline
        Function Fitting & Two Spiral & 14.93\% & 5.57\% \\
        \hline
        Function Fitting & Arb. Function & 8.10\% & 4.90\% \\
        \hline
        Image Classification & CIFAR10 & 9.32\% & 6.15\% \\
        \hline
        Image Classification & MNIST & 51.09\% & 22.69\% \\
        \hline
        Audio Classification & ESC & 10.83\% & 4.62\% \\
        \hline
        Audio Classification & Speech Cmds & 8.88\% & 0.80\% \\
        \hline
    \end{tabular}
\end{table}

\subsection{Comparisons}
Tables \ref{tab:related_regression}, \ref{tab:related_funcfit}, and \ref{tab:related_image} compare the error reductions realized in these experiments across regression, function fitting, and image classification to existing research. Comparison approaches were selected by task similarity including dataset selection as well as base compared activation function with a preference for ReLU. The audio datasets are excluded from this section as there were no relevant comparisons. Specific model improvements are noted as best case selections of each approach's performance and improvements within the most relevant scope of comparison.

It is important to note that these are not fair comparisons: important hyper-parameters such as model size, error measure, and other training specificities vary between approaches, necessarily impacting the application of trainable activation functions. Instead, these tables are included to provide a broader context for the applicability of trainable activations and a sense of where this work fits in. As well, most of the related works do not explicitly ensure that the trainable activation models are parameter equivalent to their base models, which can skew the comparison. This is noted with the PEq specifier in each table.

% Key

\begin{tablenotes}[flushleft]
    \footnotesize
    \item[] \textbf{Key:} 
            Dataset\,= Training/validation dataset;  
             Arch\,= Architecture;  
             EAF\,= Experimental Activation Function;  
             ER\,= Error reduction vs.\ ReLU;  
             EM\,= Error measure;  
             PEq\,= Parameter-equivalent comparisons;  
             SOpt\,= Separate optimizers.  
\end{tablenotes}

% -------------------- REGRESSION ----------------

\begin{table}[H]
    \centering
    \caption{Error Reduction on Regression Benchmarks}
    \label{tab:related_regression}
        \begin{tabular}{l c l c l c c c c}
            \textbf{PaperID} & \textbf{Dataset} & \textbf{Arch} &
            \textbf{EAF} & \textbf{ER} & \textbf{EM} & \textbf{PEq} & \textbf{SOpt} \\
            \hline
            
TPPLS (this) & California Housing & FNN & L-spline & 46.00\% & MSE & \checkmark & \checkmark \\
CSI* & California Housing & FNN & CR-spline & 7.27\% & NRMSE & $\times$ & $\times$ \\

            \hline            
TPPLS & Wine (red only) & FNN & L-spline & 93.87\% & MSE & \checkmark & \checkmark \\
VAF & Wine & FNN & VAF & 60.04\% & RMSE & $\times$  &  $\times$ \\
            \hline

SLAF & Boston housing & FNN & SLAF & -5.29 \% & RMSE & $\times$ & $\times$ \\

        \hline
        \end{tabular}
\end{table}

% -------------------- Function Fitting ----------------

\begin{table}[H]
    \centering
    \caption{Error Reduction on Function Fitting Benchmarks}
    \label{tab:related_funcfit}
        \begin{tabular}{l c l c l c c c c}
            \textbf{PaperID} & \textbf{Dataset} & \textbf{Arch} &
            \textbf{EAF} & \textbf{ER} & \textbf{EM} & \textbf{PEq} & \textbf{SOpt} \\
            \hline
            
SLAF & Two Spiral & FNN & SLAF & 98.40\% & ErrRate & $\times$ & $\times$ \\
LADSN & S-shape Area & FNN & B-spline & 45.95\% & ErrRate & $\times$ & $\times$ \\
LIPS & Circle Area & FNN & L-spline & 21.08\% & ErrRate & $\times$ & $\times$ \\
TPPLS (this) & Two Spiral & FNN & L-spline & 14.93\% & CrossEnt & \checkmark & \checkmark \\

        \hline
ENN & Arb. 2d function & FNN & ENN & 100.00\% & MSE & $\times$ & $\times$ \\
SLAF & Arb. 4d function & FNN & SLAF & 99.77\% & MSE & $\times$ & $\times$ \\
DiTAC & Arb. 2d function & FNN & DiTAC & 82.61\% & MSE & $\times$ & $\times$ \\
SLAF & Arb. 3d function & FNN & SLAF & 25.00\% & MSE & $\times$ & $\times$ \\
TPPLS (this) & Arb. 2d function & FNN & L-spline & 8.10\% & MSE & \checkmark & \checkmark \\

        \hline
        \end{tabular}
\end{table}

% -------------------- IMAGE CLASSIFICATION -----------------

\begin{table}[H]
    \centering
    \caption{Error Reduction on Image Classification}
    \label{tab:related_image}
        \begin{tabular}{l c l c l c c c c}
            \textbf{PaperID} & \textbf{Dataset} & \textbf{Arch} &
            \textbf{EAF} & \textbf{ER} & \textbf{EM} & \textbf{PEq} & \textbf{SOpt} \\
            \hline
VAF  & CIFAR‑10 & CNN2D & VAF & 12.59\% & ErrRate & $\times$  &  $\times$ \\
SWISH  & CIFAR‑10 & ResNet164 & Swish  & 11.29\% & ErrRate & $\times$ & $\times$ \\
SPLASH & CIFAR‑10 & LeNet‑5 & SPLASH & 10.11\% & ErrRate & $\times$ & $\times$ \\
APLU  & CIFAR‑10 & CNN2D & APLU &  9.39\% & ErrRate & $\times$ & $\times$ \\
TPPLS  (this) & CIFAR‑10 & CNN2D & L-spline &  9.32\% & CrossEnt   & \checkmark & \checkmark \\
SLAF & CIFAR-10 & FNN & SLAF & 5.19\% & ErrRate & $\times$ & $\times$ \\
LADSN  & CIFAR‑10 & ResNet32 & B‑spline &  4.60\% &  ErrRate  & $\times$ &  \checkmark \\
WCOMB  & CIFAR‑10 & CNN2D & WCOMB  &  3.88\% & ErrRate & $\times$ & $\times$ \\
LEAF  & CIFAR‑10 & CNN2D & LEAF & 3.20\% & ErrRate & $\times$ & \checkmark \\
THS    & CIFAR‑10 & ResNet34 & Tanh‑Soft‑1 &  2.55\% & ErrRate & $\times$ & $\times$ \\
APALU & CIFAR‑10 & ResNet50 & APALU &  2.40\% & ErrRate  & $\times$ & $\times$ \\
AHAF  & CIFAR‑10 & CNN2D & AHAF & -0.29\% & ErrRate  &  $\times$  & \checkmark \\
ShiLU & CIFAR‑10 & ResNet54 & ShiLU  & -100.94\% &  ErrRate & $\times$ &  $\times$ \\

            \hline
TPPLS (this) & MNIST & CNN2D      & L-spline     &   51.09\%   & CrossEnt   & \checkmark & \checkmark \\[2pt]
VAF & MNIST & CNN2D & VAF  & 33.33\%& ErrRate & $\times$ & $\times$ \\
SLAF & MNIST & FNN & SLAF & 31.82\% & ErrRate & $\times$ & $\times$ \\
THS & MNIST & CNN2D & Tanh‑soft‑1 & 30.23\%& ErrRate & $\times$ & $\times$ \\
APALU & MNIST & GoogleNet  & APALU &  9.76\%& ErrRate & $\times$ & $\times$ \\
SPLASH & MNIST & LeNet‑5 & SPLASH &  7.21\%& ErrRate & $\times$ & $\times$ \\
ShiLU & MNIST & FNN & ShiLU & 6.45\% & ErrRate & $\times$ & $\times$ \\
SplineCNN & MNIST & CNN2D & B‑spline & -16.42\%& mIoU & $\times$ & $\times$ \\

            \hline
DiTAC & ImageNet50 & ResNet34 & DiTAC & 24.00\% & ErrRate & $\times$ & $\times$ \\
DDDR & ImageNet2012 & CNN2D & PReLU & 4.37\% & ErrRate & $\times$ & $\times$ \\
            \hline
        \end{tabular}
\end{table}

\textbf{PaperID expansions:}
\\ TPPLS = Developing Training Procedures for
Piecewise-linear Spline Activation Functions in Neural
Networks, 2025 (this work).%
\\ APALU = APALU: A Trainable, Adaptive Activation Function for Deep Learning Networks, 2024 \cite{Subramanian2024}.%
\\ DiTAC = Trainable Highly-expressive Activation Functions, 2024 \cite{Chelly2024}.%
\\ ENN = ENN: A Neural Network with DCT Adaptive
Activation Functions, 2023 \cite{Martinez_Gost_2024}.%
\\ LEAF = Learnable Extended Activation Function for Deep Neural Networks, 2023 \cite{Bodyanskiy2023}.%
\\ ShiLU = Trainable Activations for Image
            Classification, 2023 \cite{Pishchik2023}.%
\\ AHAF = Adaptive Hybrid Activation Function for Deep Neural Networks, 2022 \cite{Bodyanskiy2022}.%
\\ THS = TanhSoft—Dynamic Trainable Activation Functions for Faster Learning and Better Performance, 2021 \cite{Biswas2021}.%
\\ LADSN = Learning Activation Functions in Deep (Spline) Neural Networks, 2020 \cite{Bohra2020}.%
\\ SPLASH = Splash: Learnable Activation Functions for Improving Accuracy and Adversarial Robustness, 2020 \cite{Tavakoli2020}.%
\\ SLAF = Learning Activation Functions: A new paradigm of understanding Neural Networks, 2020 \cite{goyal2020}.%
\\ LIPS = Deep Neural Networks with Trainable Activations and Controlled Lipschitz Constant, 2020 \cite{Aziznejad2020}.%
\\ VAF = A simple and eﬃcient architecture for trainable activation functions, 2019 \cite{Apicella_2019}.%
\\ WCOMB = Learning Combinations of Activation Functions, 2018 \cite{Manessi2018}.%
\\ SplineCNN = SplineCNN: Fast Geometric Deep Learning with Continuous B-Spline Kernels, 2017 \cite{Fey2018}.%
\\ SWISH = Searching for Activation Functions, 2018 \cite{Ramachandran2017}.%
\\ CSI = Learning activation functions from data, 2016 \cite{Scardapane2016}. *Compares to tanh, not ReLU.%
\\ APLU = Learning Activation Functions to Improve Deep Neural Networks, 2015 \cite{Agostinelli2015}.%
\\ DDDR = Delving Deep into Rectifiers:
            Surpassing Human-Level Performance on ImageNet Classification, 2015 \cite{he2015}.%

\section{Discussion}

\subsection{Findings}
This work serves to explore and provide reference for dual-optimization dynamics and techniques for training activation functions with a focus on linear splines. I compare trainable linear spline approaches to ReLU activation functions on an explicitly per-parameter basis. These experiments find that models with trainable linear spline activations can reliably reach lower error rates than parameter-equivalent ReLU based models. Improvements up to 94\% on regression, 15\% on function fitting, 51\% on vision, and 11\% on audio tasks were realized. The architecture change does appear to generalize across datasets and tasks, albeit variably. The spline based models also demonstrated an ability to reach low model errors that ReLU models could not, regardless of parameter count or training. Although the ReLU to isolated B-spline approach generally underperformed the others, it was highly effective on the California Housing regression dataset with immediate, minimal-cost improvements over a base model. Notably, this dataset was the most well refined and optimized as it was the basis for my senior thesis which proceeded this work.

\subsection{Limitations}
B-spline based models are more complex to implement compared to standard activations and improvements do appear to vary depending on the task and data. As well, splines are notably slower in terms of inference time, leading to end models with higher latencies. This research was completed in a compute and time constrained environment, which limited model sizes, dataset sizes and selection, as well as iteration cycles and hyperparameter tuning.

\subsection{Potential applications}
The results of this research indicate that parameterized piecewise-linear activation functions may be useful when the goal is to prioritize minimizing model error and size at the cost of additional latency and training resources. Experiments show methodologies for training more accurate models from scratch as well as for improving existing, pretrained models by swapping in spline activation functions for additional training.

\subsection{Future Works}
Future works might explore higher order B-splines, more control points, extensions to non-ReLU baseline activations, optimizations on spline latency (CUDA kernel and base implementation), as well as continued experiments on more tasks and datasets.

\section{References}
\printbibliography[heading=none]

\end{document}